\PassOptionsToPackage{unicode}{hyperref}
\PassOptionsToPackage{hyphens}{url}
\documentclass[
]{article}
\usepackage{amsmath,amssymb}
\usepackage{iftex}
\ifPDFTeX
  \usepackage[T1]{fontenc}
  \usepackage[utf8]{inputenc}
  \usepackage{textcomp} % provide euro and other symbols
\else % if luatex or xetex
  \usepackage{unicode-math} % this also loads fontspec
  \defaultfontfeatures{Scale=MatchLowercase}
  \defaultfontfeatures[\rmfamily]{Ligatures=TeX,Scale=1}
\fi
\usepackage{lmodern}
\ifPDFTeX\else
\fi
\IfFileExists{upquote.sty}{\usepackage{upquote}}{}
\IfFileExists{microtype.sty}{% use microtype if available
  \usepackage[]{microtype}
  \UseMicrotypeSet[protrusion]{basicmath} % disable protrusion for tt fonts
}{}
\makeatletter
\@ifundefined{KOMAClassName}{% if non-KOMA class
  \IfFileExists{parskip.sty}{%
    \usepackage{parskip}
  }{% else
    \setlength{\parindent}{0pt}
    \setlength{\parskip}{6pt plus 2pt minus 1pt}}
}{% if KOMA class
  \KOMAoptions{parskip=half}}
\makeatother
\usepackage{xcolor}
\providecommand{\tightlist}{%
  \setlength{\itemsep}{0pt}\setlength{\parskip}{0pt}}
\newlength{\cslhangindent}
\newlength{\csllabelwidth}
\newlength{\cslentryspacingunit} % times entry-spacing
\newenvironment{CSLReferences}[2] % #1 hanging-ident, #2 entry spacing
 {% don't indent paragraphs
  \setlength{\parindent}{0pt}
  \ifodd #1
  \let\oldpar\par
  \def\par{\hangindent=\cslhangindent\oldpar}
  \fi
  \setlength{\parskip}{#2\cslentryspacingunit}
 }%
 {}
\usepackage{calc}

\ifLuaTeX
  \usepackage{selnolig}  % disable illegal ligatures
\fi
\IfFileExists{bookmark.sty}{\usepackage{bookmark}}{\usepackage{hyperref}}
\IfFileExists{xurl.sty}{\usepackage{xurl}}{} % add URL line breaks if available
\hypersetup{
  pdftitle={YASMOT: Yet another stereo image multi-object tracker},
  pdfauthor={Ketil Malde},
  hidelinks,
  pdfcreator={LaTeX via pandoc}}

\title{YASMOT: Yet another stereo image multi-object tracker}
\author{Ketil Malde}
\date{2025-06-19}

\begin{document}
\maketitle

\hypertarget{summary}{%
\section{Summary}\label{summary}}

There now exists many popular object detectors based on deep learning
that can analyze images and extract locations and class labels for
occurrences of objects. For image time series (\emph{i.e.}, video or
sequences of stills), tracking objects over time and preserving object
identity can help to improve object detection performance, and is
necessary for many downstream tasks, including classifying and
predicting behaviors, and estimating total abundances. Here we present
\texttt{yasmot}, a lightweight and flexible object tracker that can
process the output from popular object detectors and track objects over
time from either monoscopic or stereoscopic camera configurations. In
addition, it includes functionality to generate consensus detections
from ensembles of object detectors.

\hypertarget{statement-of-need}{%
\section{Statement of need}\label{statement-of-need}}

\texttt{yasmot} is a multi-object tracker, implemented in Python, and
available under a GPLv2 license. In addition to tracking objects over
time, it can link observations between left and right cameras in a
stereo configuration, which further improves detection performance, and
allows extracting depth information and estimate the sizes of objects.
It has been tested on the output from RetinaNet (Lin et al. 2018) the
YOLO family (Redmon et al. 2016) of object detectors.

In contrast to more complex approaches that rely on analyzing the image
contents (cf.~Related Work, below), \texttt{yasmot} works with
detections only, reading observations from a separate object detector,
and linking them across time based on the relative position and
dimensions of bounding boxes and on the classification labels and
confidence scores. As a result, \texttt{yasmot} is a fast, lightweight
alternative with few dependencies.

Tracks are calculated by calculating distances between detections in two
frames, and finding an optimal pairing using the Hungarian algorithm
(Kuhn 1955). Distances are calculated by applying a Gaussian to
detection parameters (i.e., position and size coordinates) separately.
In contrast to IoU-based distances, the use of Gaussian distances allows
detections to be connected even if non-overlapping, which is important
for low frame rates and between stereo frames of objects close to the
camera. The sharpness of the Gaussian is controlled by the
\texttt{-\/-scale} parameters (see below). The Hungarian algorithm
solves assignment on a bipartite graph, so it can only work on two
frames at a time. It is possible to generalize it to consider multiple
simultaneous frames (for instance, when calculating consensus from
multiple detectors or tracking stereo images over time), but the
computational cost become prohibitive and a heuristic is used instead.

\hypertarget{usage-and-options}{%
\section{Usage and options}\label{usage-and-options}}

\hypertarget{controlling-tracking-sensitivity}{%
\subsection{Controlling tracking
sensitivity}\label{controlling-tracking-sensitivity}}

The \texttt{-\/-scale} parameter controls how the different bounding box
pairs are ranked when considered for tracking (or stereo matching). The
algorithm uses a Gaussian score for position and size, and this
parameter controls the sharpness (or temperature) of the Gaussian.
Generally, if you have large changes between frames (rapidly moving
objects or low frame rate, you can try reducing this parameter.

\hypertarget{handling-missing-detections}{%
\subsection{Handling missing
detections}\label{handling-missing-detections}}

Object detectors are sometimes unreliable, and objects may fail to be
detected, especially when they are partially occluded or otherwise
difficult to identify. To remedy this, \texttt{yasmot} will attempt to
link tracks across missing detections. The maximum gap allowed in a
track is controlled by the parameter \texttt{-\/-max\_age}. The age is
derived from the frame name, and unless the frame name is a plain
number, the extraction can be specified with \texttt{-\/-time\_pattern}.

If desired, the missing observations can be inferred by specifying the
\texttt{-\/-interpolate} option. The interpolated detections can be
identified from their probability, which is set to 0.0000.

In case there are classes representing an unknown or unidentified
object, it is possible to specify the label with the
\texttt{-\/-unknown} parameter to avoid having this class be called as a
consensus class for a track.

\hypertarget{tracking-stereo-images}{%
\subsection{Tracking stereo images}\label{tracking-stereo-images}}

The \texttt{-s} option links objects taken with a stereoscopic camera
setup. Normally tracks will be generated, but the \texttt{-\/-no-track}
option can be specified to only link detections between the cameras, and
not in time.

\hypertarget{ensemble-predictions}{%
\subsection{Ensemble predictions}\label{ensemble-predictions}}

If you run multiple object detectors, it may be useful to combine the
outputs into a consensus set of predictions. This can be achieved by
specifying the \texttt{-c} option. Again you can use
\texttt{-\/-no-track} if you just want the frame-by-frame consensus and
not perform tracking as well.

\hypertarget{using-pixel-based-coordinates}{%
\subsection{Using pixel-based
coordinates}\label{using-pixel-based-coordinates}}

The YOLO object detector (Redmon et al. 2016) outputs image coordinates
as fractional images, i.e.~values in the range from 0 to 1. Other object
detectors, like RetinaNet (Lin et al. 2018), may output a CSV file with
pixel-based coordinates. Since \texttt{yasmot} does not require the
images to be available, you therefore may need to specify the pixel size
of the images, e.g.~as \texttt{-\/-shape\ 1228,1027} when using
pixel-based coordinates.

\hypertarget{output-formats}{%
\subsection{Output formats}\label{output-formats}}

If the \texttt{-o} option is not specified, the output will be written
to \texttt{stdout}. If the option is specified as \texttt{-o\ outfile},
the following files may result, depending on the selection of other
parameters specified:

\begin{itemize}
\tightlist
\item
  \texttt{outfile.frames} - frame annotations in YOLO format, with track
  number added. For stereo images, each line will contain the bounding
  boxes for both images.
\item
  \texttt{outfile.tracks} - the list of tracks with the sequence of
  observations that constitute each track.
\item
  \texttt{outfile.pred} - per track consensus class predictions.
\end{itemize}

\hypertarget{examples}{%
\section{Examples}\label{examples}}

The following examples are taken from the included test suite and the
data files can be found in the \texttt{tests} directory.

\textbf{Perform tracking on a directory of predictions from YOLO}

This reads a directory with a text file of annotations for each frame:

\begin{verbatim}
yasmot tests/lab2
\end{verbatim}

\textbf{Perform tracking with interpolation}

Interpolation creates virtual annotations to fill in gaps (i.e., missing
detections) in the tracks:

\begin{verbatim}
yasmot tests/lab2 --interpolate
\end{verbatim}

\textbf{Perform tracking limiting gap size}

Here, we only connect tracks with a maximum of two frames without
detections. The frames in \texttt{tests/lab} are named
\texttt{frame\_000152.txt}, \texttt{frame\_000153.txt}, and so on, and
the \texttt{-\/-time\_pattern} expression must match this format.

\begin{verbatim}
yasmot --max_age 2 --time_pattern frame_\{:d\}.txt tests/lab2
\end{verbatim}

\textbf{Perform tracking on stereo images}

The \texttt{-s} option requires the user to specify two input stream,
with the assumption that the left images are specified before the right
images. Here we process predictions in pixel-based CSV format, and thus
we must specify the images size with \texttt{-\/-shape}:

\begin{verbatim}
yasmot -s --shape 1228,1027 tests/stereo1_Left.csv tests/stereo1_Right.csv
\end{verbatim}

\textbf{Only link stereo predictions without tracking}

Again we use pixel-based RetinaNet predictions between the two cameras:

\begin{verbatim}
yasmot -s --no-track --shape 1228,1027 tests/stereo1_Left.csv tests/stereo1_Right.csv
\end{verbatim}

\textbf{Merge predictions from multiple object detectors}

Here we use predictions from a family of YOLO v8 models to provide
ensemble predictions:

\begin{verbatim}
yasmot -c tests/consensus/y8x*
\end{verbatim}

\hypertarget{related-work}{%
\section{Related work}\label{related-work}}

Tracking objects has long been recognized as a fundamental task in
computer vision, with applications ranging from surveillance to
autonomous systems. The widely-used image processing library OpenCV has
incorporated several algorithms and components dedicated to object
tracking, reflecting the task's importance. These include BOOSTING, MIL,
KCF, CSRT, MedianFlow, TLD, MOSSE, and GOTURN (Bradski 2000). Each of
these methods offers distinct approaches to balancing speed, accuracy,
and robustness, catering to a variety of real-time tracking needs.
Another commonly used tool is SORT (Simple Online and Realtime Tracking)
(Bewley et al. 2016), which uses uses a Kalman filter to predict object
motion and associates predictions with detections using Intersection
over Union (IoU).

The advent of deep learning object detectors like YOLO have brought new
object tracking tools to the fore, and the popular implementations of
YOLO by Ultralytics (Jocher, Chaurasia, and Qiu 2023) integrate two such
tracking algorithms, ByteTrack and BoT-SORT. Like yasmot, ByteTrack
processes object detection model output to associate bounding boxes
across frames, but it uses intersection over union (IoU) instead of
Gaussian distances to link detections. BoT-SORT (Aharon, Orfaig, and
Bobrovsky 2022), on the other hand, extends SORT by incorporating
additional motion and appearance cues to enhance tracking precision.

Other object trackers that examine the detected objects to support
tracking Tracktor++ (Bergmann, Meinhardt, and Leal-Taixe 2019), and
DeepSORT (Wojke, Bewley, and Paulus 2017), which similarly to BoT-SORT
extends SORT with features from a deep learning model to matches
detections across frames more reliably, particularly in crowded or
dynamic environments.

\hypertarget{availability}{%
\section{Availability}\label{availability}}

The program is available via PyPI (as \texttt{pip\ install\ yasmot}) or
from GitHub \url{https://github.com/ketil-malde/yasmot}.

\hypertarget{acknowledgments}{%
\section{Acknowledgments}\label{acknowledgments}}

This work was developed using data from the CoastVision (RCN 325862) and
CRIMAC projects (RCN 309512), and after productive discussions with
Vaneeda Allken, Taraneh Westergerling, and Peter Liessem.

\hypertarget{references}{%
\section*{References}\label{references}}
\addcontentsline{toc}{section}{References}

\hypertarget{refs}{}
\begin{CSLReferences}{1}{0}
\leavevmode\vadjust pre{\hypertarget{ref-aharon2022bot}{}}%
Aharon, Nir, Roy Orfaig, and Ben-Zion Bobrovsky. 2022. {``{BoT-SORT}:
Robust Associations Multi-Pedestrian Tracking.''} \emph{arXiv Preprint
arXiv:2206.14651}.

\leavevmode\vadjust pre{\hypertarget{ref-bergmann2019tracking}{}}%
Bergmann, Philipp, Tim Meinhardt, and Laura Leal-Taixe. 2019.
{``Tracking Without Bells and Whistles.''} In \emph{Proceedings of the
IEEE/CVF International Conference on Computer Vision}, 941--51.

\leavevmode\vadjust pre{\hypertarget{ref-bewley2016simple}{}}%
Bewley, Alex, Zongyuan Ge, Lionel Ott, Fabio Ramos, and Ben Upcroft.
2016. {``Simple Online and Realtime Tracking.''} In \emph{2016 IEEE
International Conference on Image Processing (ICIP)}, 3464--68. IEEE.

\leavevmode\vadjust pre{\hypertarget{ref-opencv_library}{}}%
Bradski, G. 2000. {``{The OpenCV Library}.''} \emph{Dr. Dobb's Journal
of Software Tools}.

\leavevmode\vadjust pre{\hypertarget{ref-yolov8_ultralytics}{}}%
Jocher, Glenn, Ayush Chaurasia, and Jing Qiu. 2023. {``Ultralytics
YOLOv8.''} \url{https://github.com/ultralytics/ultralytics}.

\leavevmode\vadjust pre{\hypertarget{ref-kuhn1955hungarian}{}}%
Kuhn, Harold W. 1955. {``The Hungarian Method for the Assignment
Problem.''} \emph{Naval Research Logistics Quarterly} 2 (1-2): 83--97.

\leavevmode\vadjust pre{\hypertarget{ref-lin2018focallossdenseobject}{}}%
Lin, Tsung-Yi, Priya Goyal, Ross Girshick, Kaiming He, and Piotr Dollár.
2018. {``Focal Loss for Dense Object Detection.''}
\url{https://arxiv.org/abs/1708.02002}.

\leavevmode\vadjust pre{\hypertarget{ref-Redmon_2016_CVPR}{}}%
Redmon, Joseph, Santosh Divvala, Ross Girshick, and Ali Farhadi. 2016.
{``You Only Look Once: Unified, Real-Time Object Detection.''} In
\emph{Proceedings of the IEEE Conference on Computer Vision and Pattern
Recognition (CVPR)}.

\leavevmode\vadjust pre{\hypertarget{ref-wojke2017simple}{}}%
Wojke, Nicolai, Alex Bewley, and Dietrich Paulus. 2017. {``Simple Online
and Realtime Tracking with a Deep Association Metric.''} In \emph{2017
IEEE International Conference on Image Processing (ICIP)}, 3645--49.
IEEE.

\end{CSLReferences}

\end{document}